\documentclass[sigconf]{acmart}
\AtBeginDocument{%
  }

\copyrightyear{2026}
\acmYear{2026}
\setcopyright{cc}
\setcctype{by}
\acmConference[GeoSim '26]{The 9th ACM SIGSPATIAL International Workshop on GeoSpatial Simulation}{November 03--06, 2026}{Riverside, CA, USA}
\acmBooktitle{The 9th ACM SIGSPATIAL International Workshop on GeoSpatial Simulation (GeoSim '26), November 03--06, 2026, Riverside, CA, USA}
\acmDOI{10.1145/3849737.3849831}
\acmISBN{979-8-4007-3059-7/2026/11}

\usepackage{multirow}
\begin{document}

\title[Editable Map-Conditioned Trajectory Generation]{Editable Map-Conditioned Trajectory Generation for Human Mobility Simulation}


\author{Takayuki Mizuno}
\authornote{Both authors contributed equally to this research.}
\authornote{Corresponding author.}
\orcid{0000-0003-1332-076X}
\affiliation{%
 \institution{National Institute of Informatics}
 \city{Tokyo}
 \country{Japan}}
\email{mizuno@nii.ac.jp}

\author{Shouji Fujimoto}
\authornotemark[1]
\orcid{0000-0001-7195-634X}
\affiliation{%
  \institution{Kanazawa Gakuin University}
  \city{Kanazawa}
  \state{Ishikawa}
  \country{Japan}
}
\email{fujimoto@kanazawa-gu.ac.jp}

\author{Mikito Hiruki}
\affiliation{%
  \institution{The University of Tokyo}
  \city{Tokyo}
  \country{Japan}}
\email{hiruki@g.ecc.u-tokyo.ac.jp}

\author{Atushi Ishikawa}
\affiliation{%
  \institution{Kanazawa Gakuin University}
  \city{Kanazawa}
  \state{Ishikawa}
  \country{Japan}
}
\email{ishikawa@kanazawa-gu.ac.jp}

\renewcommand{\shortauthors}{Mizuno et al.}

\begin{abstract}
Geospatial simulation of infrastructure interventions requires mobility generators that respond directly to edited maps, yet many data-driven generators do not expose the map as an editable condition. We formulate this task as map-conditioned autoregressive generation of human mobility: a road raster conditions a decoder that emits nominal $31.25\,\mathrm{m}$ mesh-cell tokens at one-minute intervals. The mesh-local vocabulary supports held-out and locally edited maps without retraining or vocabulary changes. We instantiate a ResNet-50 visual-prefix configuration and a Vision Transformer (ViT) cross-attention configuration, trained from scratch on $87{,}400$ smartphone-derived trajectories from $874$ meshes in Ishikawa Prefecture, Japan; $219$ meshes are held out. We evaluate map sensitivity by comparing correct-map and within-split shuffled-map generations with held-out real trajectories. On the $110$-mesh test split, for the ResNet-50 configuration, correct-map generations are closer than shuffled-map generations on $60\%$ of meshes under Hausdorff-based energy distance ($p=0.021$), while DTW is directional but inconclusive ($57\%$, $p=0.074$); correlation with real density is $0.38$ with the correct map versus $0.01$ with shuffled maps. The ViT configuration shows weaker trajectory-level sensitivity and smaller density gains. An illustrative bridge-removal edit changes generated continuations without retraining. Together, these results support the feasibility of editable-map human-mobility simulation.
\end{abstract}




\maketitle

\section{Introduction}
\label{sec:introduction}

How would human movement change if a bridge were closed, a street were added, or a district layout were redesigned? Mobility models have informed policy interventions at broader scales, including reopening decisions during COVID-19~\cite{Chang2021}; here we focus on localized changes to the built environment. Such interventions are difficult for learned mobility generators because changed infrastructure is often not exposed as a directly editable input. A mobility digital twin needs both a representation of the proposed environment and a generator that can produce movement under that representation~\cite{Wang2022}. We therefore study a concrete interface question: can an autoregressive model take an editable road-map raster and generate anonymized human-mobility trajectories that respond to unseen or locally modified road geometry? The data used here are smartphone-derived movements across travel modes; they are not restricted to pedestrians, cars, or public transport.

Existing approaches offer complementary strengths. The geosimulation tradition of automata-based urban modeling~\cite{Benenson2004} established the map as the substrate on which movement and land-use change are simulated; our work can be read as an update of this idea for learned, autoregressive generators. Rule-based simulators such as MATSim~\cite{Horni2016} and SUMO~\cite{Lopez2018} provide explicit infrastructure and intervention models, but realistic heterogeneous behavior requires substantial specification and calibration. Data-driven trajectory models instead learn sequences of coordinates, grid cells, venues, or road segments using recurrent, attention-based, imitation-learning, or generative architectures~\cite{Feng2018,Luo2021,Yabe2024,Luca2021,Mizuno2026,Choi2021}. They can reproduce movement patterns in a fixed spatial representation, but a map edit is usually handled indirectly through an updated graph, vocabulary, constraint set, or training corpus. Diffusion models increasingly incorporate road-network topology and spatio-temporal constraints~\cite{Zhu2023,Zhu2024,Wei2024,Zhu2025}. Our aim is not to replace these approaches or to propose a new image encoder; it is to expose the road map itself as an editable conditioning interface while retaining stepwise autoregressive continuation.

Figure~\ref{fig:task} gives the concrete input and output. For one cell of the Japanese standard regional grid (nominally $1\,\mathrm{km}\times1\,\mathrm{km}$), the input is a $224\times224$ road raster. The output is a sequence of discrete position tokens on a mesh-local $32\times32$ grid; each token denotes a nominal $31.25\,\mathrm{m}$ cell and corresponds to one minute in the source trajectories. The target trajectory is not part of the input image. The red path in Figure~\ref{fig:task} is drawn only after generation, by converting emitted tokens to geographic cell centroids. Because token identities are mesh-local, the same decoder and vocabulary can be applied to a held-out mesh or an edited raster without introducing new absolute-location identifiers.

This paper makes three contributions. First, it formulates editable-map human-mobility simulation as image-conditioned autoregressive sequence generation. Its contribution is the interface: an editable raster conditions stepwise spatial-token decoding, while the CNN and ViT are standard components. Second, it introduces a matched-versus-shuffled map protocol that tests whether generated distributions are sensitive to the specific road geometry. Both conditions are compared with held-out real trajectories using trajectory-level energy distance and density-level heatmap correlation. Third, it evaluates $1{,}093$ meshes in Ishikawa Prefecture, Japan, with $87{,}400$ sampled training trajectories ($699{,}200$ augmented pairs), and illustrates a bridge removal without retraining. This is a feasibility evaluation; external baselines, absolute-realism metrics, and quantitative intervention validation remain future work.

\section{Related Work}
\label{sec:related}

We organize related work around four design choices relevant to this interface: how trajectories are represented, how maps condition generation, how positions are tokenized, and how visual encoders condition autoregressive decoders.

\subsection{Coordinate-based Trajectory Prediction Models}
\label{sec:related:coord}

The dominant paradigm in data-driven mobility modeling represents a trajectory as a sequence of discrete location tokens---for example, latitude--longitude bins, grid indices, venue IDs, or road-segment IDs---and learns transition dynamics over this vocabulary. Recurrent architectures with history-aware attention~\cite{Feng2018}, self-attention over non-contiguous check-ins~\cite{Luo2021}, and geography-aware location embeddings~\cite{Lian2020} have progressively sharpened next-location prediction. Casting trajectory generation as language modeling was introduced by Mizuno et al.~\cite{Mizuno2022}, who trained a GPT-2 decoder autoregressively on urban daily trajectories; the paradigm has since been extended to larger corpora and stronger benchmarks~\cite{Mizuno2026,Yabe2024,Luca2021}.

These models are not inherently incapable of representing new infrastructure; the limitation for our use case is the intervention interface. A road edit is not expressed simply by changing an input image: the graph, location set, embeddings, constraints, or training data typically must be updated. Our formulation targets a complementary capability---preserving an autoregressive trajectory state while exposing local road geometry as a directly editable visual condition.

\subsection{Map-conditioned Generative Models}
\label{sec:related:map}

A second, more recent line of work addresses the geometric blindness of purely coordinate-based predictors by conditioning trajectory generation directly on spatial context, often producing a complete trajectory as a high-dimensional sample from a distribution conditioned on a map, road network, or constraint representation. DiffTraj~\cite{Zhu2023} realizes this paradigm with a denoising diffusion model over continuous GPS coordinates; ControlTraj~\cite{Zhu2024} incorporates road-network topology as a structural constraint within the diffusion process; and Diff-RNTraj~\cite{Wei2024} refines the output representation by combining discrete road segments with a continuous travel-rate variable. A recent survey~\cite{Zhu2025} situates this rapidly growing family within the broader landscape of unconditional and conditional trajectory generators. Adjacent to this diffusion-based lineage, TrajGAIL~\cite{Choi2021} formulates on-network trajectory generation as imitation learning over a pre-discretized road-segment vocabulary. These models are important points of comparison at the level of problem motivation, but their input/output interfaces differ from the one studied here: they are designed primarily to generate complete trajectories under a fixed spatial representation, whereas our experiment asks whether an editable raster map can condition an autoregressive decoder at each step.

These models are relevant comparators for absolute trajectory realism and should be included in a future common benchmark. Their interfaces, however, differ from the narrow question studied here: many sample a complete trajectory under a fixed representation, whereas we ask whether an editable raster affects stepwise continuation under the same decoder. Accordingly, the present evaluation isolates map sensitivity through matched-versus-shuffled conditioning and encoder configurations. This design does not establish superiority to existing generators; it tests whether the proposed simulator responds to the specific map rather than to generic road-image statistics.

\subsection{Spatial Tokenization for Trajectory Generation}
\label{sec:related:token}

Discrete spatial tokens connect raster pixels with autoregressive sequence generation~\cite{Mai2022}. We use a mesh-local $32\times32$ grid so that each emitted position token indexes a cell of the conditioning raster rather than an absolute coordinate. Latitude--longitude values are recovered only after generation by mapping each token to its cell centroid. Section~\ref{sec:method:token} defines the vocabulary, special token, and nominal cell size.

\subsection{Image Captioning and Vision-Language Models}
\label{sec:related:caption}

Image captioning provides the methodological template on which our formulation is modeled: a visual encoder compresses an image into an embedding, and an autoregressive decoder emits a sentence conditioned on it. The Show and Tell architecture of Vinyals et al.~\cite{Vinyals2015} established this encoder--decoder paradigm end-to-end with a convolutional encoder and a recurrent decoder; subsequent work replaced the language side with Transformer decoders and the visual side with Vision Transformer backbones, yielding fully Transformer-based captioners~\cite{Fang2022}. A complementary line reduces the burden on the visual encoder by reusing pretrained language models: ClipCap~\cite{Mokady2021} projects a CLIP image embedding through a lightweight mapping network into a \emph{visual prefix} of continuous pseudo-tokens that steer a frozen GPT-2 decoder.

Two architectural motifs from this literature structure our design. The \emph{visual prefix}, adopted for our CNN-based encoder, prepends encoder-produced tokens to the decoder input so that each subsequent prediction is conditioned on the image through causal self-attention. \emph{Cross-attention conditioning}, used for our ViT-based encoder, exposes patch tokens to the decoder through dedicated cross-attention layers. Their juxtaposition lets us test how two standard visual-conditioning designs with different local priors behave in the same trajectory task.

The transfer from captioning is operational: the map raster plays the role of the input image, and the spatial-token trajectory plays the role of the caption. Here, a \emph{conditioning image} simply means the raster supplied to the visual encoder; it does not contain the target path. The output is not a semantic description but an ordered movement sequence whose tokens are decoded to mesh-cell centroids.

\section{Methodology}
\subsection{Problem Formulation}
\label{sec:method:problem}

We model a trajectory as an autoregressive sequence conditioned on a map raster. Let $m$ denote a third-level cell of the Japanese standard regional grid, nominally $1\,\mathrm{km}\times1\,\mathrm{km}$; its physical dimensions vary slightly with latitude. Let $I_m \in \mathbb{R}^{H \times W \times C}$ be either an aggregate mobility-density raster used for training or an editable OSM-derived road raster used for validation, testing, and intervention (Section~\ref{sec:experiments:maps}). The target path is never drawn on $I_m$. A trajectory $s=(s_1,\ldots,s_T)$ contains one position token per one-minute sample followed by a period token, where $T$ is the total sequence length. The nominal vocabulary has $1{,}030$ entries: $1{,}024$ mesh-cell tokens, one period token, and five reserved entries (Section~\ref{sec:method:token}). The model factorizes the conditional distribution autoregressively:
\begin{equation}
  p_\theta\!\left(s \,\big|\, I_m,\, s_{1:k}\right)
  \;=\; \prod_{t = k+1}^{T}
  p_\theta\!\left(s_t \,\big|\, s_{<t},\, I_m\right),
  \label{eq:ar}
\end{equation}
where $s_{1:k}$ is an optional observed prefix and the model predicts $s_{k+1:T}$. We use $k=0$ for map-only generation and $k=2$ for trajectory-level comparison and the bridge intervention. The parameter set $\theta$ collects the image encoder and decoder-only Transformer, all learned from scratch without pretrained language-model or vision--language weights.

\begin{figure}[t]
  \centering
  \includegraphics[width=\linewidth]{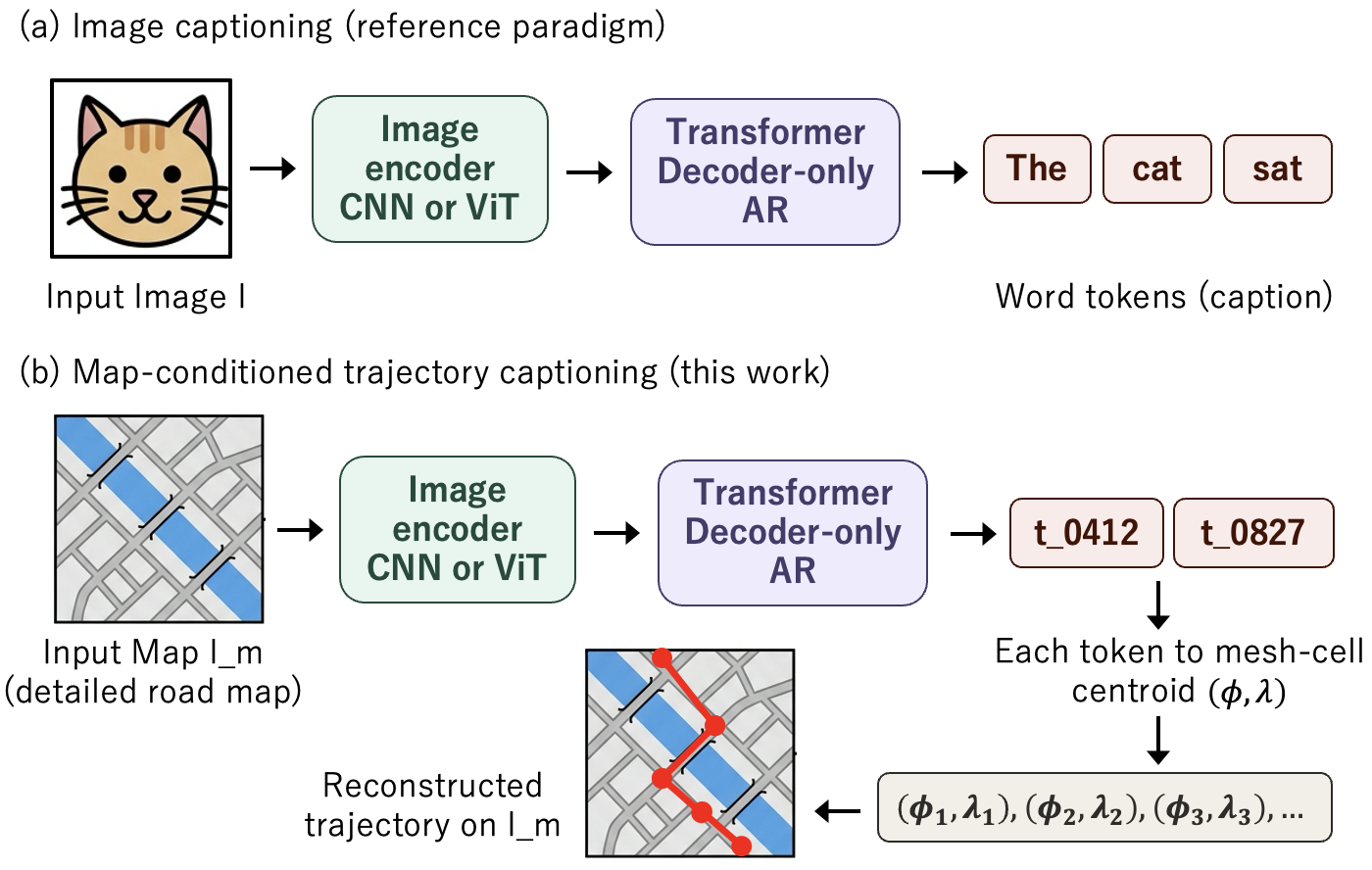}
  \caption{Overview of the input and output. Top: standard image captioning maps an image to word tokens. Bottom: the proposed model maps a road raster $I_m$ to mesh-cell tokens. The red trajectory is not part of the input; it is overlaid only after the generated tokens are converted to geographic cell centroids. All parameters are learned from scratch.}
  \Description{A schematic compares standard image captioning with the proposed map-conditioned trajectory generator. A road raster is encoded and decoded into mesh-cell tokens, which are converted to geographic centroids and plotted as a trajectory.}
  \label{fig:task}
\end{figure}

For a position token, a deterministic map
$\tau:\mathcal{V}_{\mathrm{pos}}\rightarrow\mathbb{R}^2$
returns the geographic centroid of its cell:
\begin{equation}
  \tau(s_t) \;=\; (\varphi_t,\, \lambda_t).
  \label{eq:decode}
\end{equation}
The generated trajectory is the ordered list of these centroids after the period and reserved tokens are removed. Figure~\ref{fig:task} illustrates the input--output relationship.

This factorization supports two practical modes. In map-only generation, the model samples the first position from the raster and continues step by step. In prefix-conditioned generation, it continues from observed initial positions. Because the raster conditions every predicted token, the same prefix and model can be rerun under an edited raster; the bridge example compares such original-map and edited-map continuations. We do not change the map during a single decoded trajectory in the present experiments. Mesh-local token identities also allow the same output vocabulary to be used on map geometries not seen during training.

\subsection{Spatial Tokenization}
\label{sec:method:token}

Each map image covers one third-level Japanese standard regional mesh. We subdivide it into a regular $32\times32$ grid and assign one token to each cell, giving $|\mathcal{V}_{\mathrm{pos}}|=1{,}024$. The nominal cell width is $31.25\,\mathrm{m}$; because the standard mesh is defined in latitude--longitude, its exact physical size varies slightly by location. Tokens are indexed in row-major order from the northwest corner, and $\tau$ maps each token to the geographic centroid of its cell. The indexing is \emph{mesh-local}: the same token denotes the same relative cell in every mesh, not the same absolute latitude--longitude coordinate.

A period token marks the end of a trajectory. Thus $1{,}025$ entries are active ($1{,}024$ positions plus the period), while the decoder table has $1{,}030$ rows and reserves five unused entries. The period and reserved entries are removed before coordinate reconstruction, so a length-$T$ sequence yields at most $T-1$ geographic positions.

\subsection{Map Image Encoding}
\label{sec:method:encoder}

The image encoder transforms a map image $I_m$ into a sequence of visual tokens that condition the decoder-only Transformer of Section~\ref{sec:method:decoder}. We instantiate it in two forms---a convolutional backbone and a Vision Transformer backbone---while keeping the data, tokenization, decoder size, loss, and sampling rule fixed. Because the standard CNN and ViT variants use different conditioning pathways (visual prefix versus cross-attention), the comparison should be read as an encoder--conditioning comparison rather than as an isolated backbone ablation. In both cases the input is a $224 \times 224$ road raster replicated to RGB channels and the output is a sequence of $d$-dimensional visual tokens with $d = 768$, matching the hidden size of the decoder.

\subsubsection{CNN-based encoder.}
\label{sec:method:encoder:cnn}

The convolutional encoder is a ResNet-50 backbone~\cite{He2016} from which we discard the classification head and the terminal global average pooling. We replace the latter with an adaptive average pooling layer that maps the final $2048$-channel feature tensor to a $7 \times 7$ spatial grid, preserving locality. A $1 \times 1$ convolution then linearly projects each grid cell from $2048$ to $768$ channels, yielding $7 \times 7 = 49$ visual tokens of dimension $768$. These $49$ tokens are flattened in row-major order and prepended to the spatial-token sequence as a \emph{visual prefix}, so that conditioning on the image is effected entirely through the decoder's causal self-attention, without architectural modification of the decoder. This pattern follows the visual-prefix family of captioning architectures~\cite{Mokady2021}, with the prefix here originating from a convolutional feature map rather than a CLIP-style embedding.

\subsubsection{ViT-based encoder.}
\label{sec:method:encoder:vit}

The Vision Transformer encoder~\cite{Dosovitskiy2021} partitions the $224 \times 224$ input into non-overlapping $16 \times 16$ patches, producing a $14 \times 14 = 196$ patch grid; each patch is linearly embedded and processed by a stack of self-attention blocks to yield $196$ patch tokens of dimension $768$. The encoder follows the ViT-Base configuration~\cite{Dosovitskiy2021}: $12$ Transformer layers, $12$ attention heads, feed-forward size $3{,}072$, and hidden size $768$. The patch-token sequence is exposed to the decoder through dedicated cross-attention layers inserted between the self-attention and feed-forward sublayers of each decoder block, following the canonical encoder--decoder template of fully Transformer-based captioners~\cite{Fang2022}.

\subsubsection{Rationale for the comparison.}
\label{sec:method:encoder:compare}

The two encoders carry fundamentally different inductive biases. The CNN builds its representation through hierarchical local receptive fields---edge orientations, junction shapes, segment widths---so its $7 \times 7$ output is biased toward the local geometry of the road network. The ViT applies self-attention globally from the first layer, yielding a representation biased toward long-range spatial dependencies but endowed with comparatively weak local structural priors, a contrast made explicit by representation-similarity analyses of the two families on natural images~\cite{Raglu2021}. Because next-step movement may depend strongly on nearby road geometry, the two configurations may be differently aligned with the task. However, encoder family and conditioning pathway are coupled in our implementations; Section~\ref{sec:results} therefore reports configuration-level differences rather than isolating their cause.

\subsection{Autoregressive Trajectory Decoder}
\label{sec:method:decoder}

The decoder realizes $p_\theta(s_t \mid s_{<t}, I_m)$ of Equation~\ref{eq:ar} as a decoder-only Transformer with causal self-attention~\cite{Mizuno2026}. It is a stack of $L = 12$ Transformer blocks, each comprising a causal multi-head self-attention sublayer with $h = 12$ heads and a position-wise feed-forward sublayer, with pre-layer normalization and residual connections. The hidden size is $d = 768$, the per-head dimension is $d/h = 64$, and the feed-forward intermediate dimension is $4d = 3{,}072$. The input embedding table has $|\mathcal{V}| = 1{,}030$ rows of dimension $d$, learned from scratch, and learned absolute positional embeddings are added to the token embeddings before the first block.

The two encoder variants couple to the decoder through the pathways introduced in Section~\ref{sec:method:encoder}. The convolutional backbone's $49$ visual tokens are prepended to the trajectory token stream as a length-$(49 + T)$ concatenated sequence; the causal mask allows every trajectory position to attend to every visual-prefix position while preserving strict left-to-right masking among trajectory positions. The Vision Transformer's $196$ patch tokens are instead exposed through cross-attention sublayers inserted between the self-attention and feed-forward sublayers of each decoder block.

Given a training pair $(I_m, s)$, all parameters $\theta$ are optimized jointly by minimizing the token-level negative log-likelihood:
\begin{equation}
  \mathcal{L}(\theta)
  \;=\; -\frac{1}{T - k}
  \sum_{t = k+1}^{T}
  \log p_\theta\!\left(s_t \,\big|\, s_{<t},\, I_m\right),
  \label{eq:nll}
\end{equation}
with $k=0$ during training, so the loss covers all non-padding trajectory tokens, including the period token. Teacher forcing is used throughout. At inference, the same decoder supports map-only generation with no position-token prefix and prefix-conditioned continuation from observed initial positions; in both cases, tokens are sampled autoregressively until the period token or maximum length is reached.

\section{Experimental Setup}
\label{sec:experiments}

\subsection{Dataset}
\label{sec:experiments:data}

We evaluate the framework on anonymized human-mobility records for Ishikawa Prefecture, Japan ($4{,}191\,\mathrm{km}^2$, resident population $\approx1{,}086{,}000$), provided by Agoop Corporation's Dynamic Population Data service~\cite{Agoop2026}. Each raw observation has the schema (daily anonymized device identifier, timestamp, latitude, longitude) at one-minute resolution. The data contain no transport-mode, trip-purpose, or semantic-place labels; the generated trajectories therefore represent undifferentiated human movement across travel modes and contain only ordered positions at one-minute intervals. We use July and August 2023 and January 2024 to include summer and winter movement regimes.

Ishikawa Prefecture contains $4{,}517$ third-level cells in Japan's standard regional grid. Each is nominally $1\,\mathrm{km}\times1\,\mathrm{km}$, although its exact physical dimensions vary with latitude; below, ``1-km mesh'' is shorthand for this standard unit. A trajectory is retained only if it visits at least five distinct nominal $31.25\,\mathrm{m}$ cells, excluding observations that remain stationary throughout the window. We also exclude meshes with fewer than $100$ retained trajectories. These filters leave $1{,}093$ meshes for modeling.

The $1{,}093$ modeling meshes are partitioned uniformly at random into training, validation, and test splits in the ratio $8 : 1 : 1$, yielding $874$, $109$, and $110$ meshes, respectively. The split is \emph{mesh-level} rather than trajectory-level: trajectory-level splitting would place trajectories from the same physical location in both training and test sets, conflating within-location generalization with the cross-location, map-conditioned generalization our framework is designed to address. Mesh-level splitting ensures that every test-set trajectory occurs within a mesh whose map image was never presented during training, so that test-time performance directly reflects the model's ability to generate plausible movement from the \emph{visual} content of a previously unseen map rather than from memorized location statistics.

\subsection{Map Image Construction}
\label{sec:experiments:maps}

Each of the $1{,}093$ meshes has one $224\times224$ grayscale conditioning raster. Training uses an aggregate \emph{mobility-density image} derived from observed samples. Validation, testing, and intervention use an \emph{OSM-derived road image} that can be edited directly. The first image supplies empirical road-use patterns during learning; the second retains only road geometry at generation time. No target trajectory is drawn on either image.

\subsubsection*{Mobility-density image (training).}

For every modeling mesh, the one-minute location samples over the three-month window of Section~\ref{sec:experiments:data} are binned onto a $224 \times 224$ grid that tiles the $1\,\mathrm{km}$ mesh uniformly. Let $N(u, v)$ denote the sample count at pixel $(u, v)$; the pixel intensity is
\begin{equation}
  y_{\mathrm{train}}(u, v)
  \;=\;
  \frac{\log\!\left(1 + N(u, v)\right)}
       {\max_{u', v'}\log\!\left(1 + N(u', v')\right)},
  \label{eq:heatmap}
\end{equation}
normalized to $[0,1]$ per mesh. Log scaling prevents a few high-count pixels from dominating the image. Because many observed movements occur along streets and adjacent spaces, bright pixels form road-like corridors, but the image is not an exact road mask: it may also include sidewalks, building entrances, and nearby lots. It contains only three-month aggregate counts, with no temporal order, destination label, or individual continuation. This image is used only for training. At validation, test, and edited-map generation, no mobility-density image from the target mesh is supplied.

\subsubsection*{OSM-derived road image (generation).}

At generation time, the model must be able to condition on maps for which mobility data are unavailable or intentionally perturbed---the latter being the proof-of-concept use case tested here. We therefore construct a purely geometry-based road image from OpenStreetMap~\cite{Haklay2008}. For each road category $h$ in the OSM \texttt{highway} tag, we define a binary indicator $x_h(u, v) \in \{0, 1\}$ that is one at pixels lying on a class-$h$ road centerline dilated to a category-specific width (Table~\ref{tab:osm-weights}). Each category carries a scalar brightness weight $w_h \in \mathbb{R}_{\ge 0}$, and the generation-time image is
\begin{equation}
  y_{\mathrm{gen}}(u, v)
  \;=\;
  255 \sum_{h} x_h(u, v)\, w_h.
  \label{eq:osm}
\end{equation}
The weights $\{w_h\}$ are \emph{estimated}, not hand-set: we minimize the $L_2$ loss between $y_{\mathrm{gen}}$ and $y_{\mathrm{train}}$ over all training meshes. Table~\ref{tab:osm-weights} reports the resulting values---motorways, trunks, and primary roads receive the highest weights, as expected. This regression transfers the visual appearance of the empirical road-use raster into a purely geometry-based, editable OSM representation. It reduces the train--generation visual gap while preserving the key counterfactual property: after the weights have been fitted once, held-out or edited generation uses only OSM road geometry; prefix-conditioned continuation additionally supplies initial positions, but never future mobility observations from the target map.

\begin{table}[t]
  \centering
  \small
  \caption{OpenStreetMap \texttt{highway} categories with their rendered road widths and brightness weights $w_h$. Weights are estimated by $L_2$ regression against the training mobility-density images of Equation~\ref{eq:heatmap}; categories are grouped into arterial (L3), major local (L2), residential (L1), and auxiliary tiers.}
  \label{tab:osm-weights}
  \begin{tabular}{llcc}
    \toprule
    Tier & Category & Width (m) & Weight $w_h$ \\
    \midrule
    \multirow{6}{*}{Arterial (L3)}
      & motorway       & 20 & 0.45 \\
      & motorway\_link & 20 & 0.16 \\
      & trunk          & 20 & 0.41 \\
      & trunk\_link    & 20 & 0.19 \\
      & primary        & 20 & 0.37 \\
      & primary\_link  & 20 & 0.16 \\
    \midrule
    \multirow{4}{*}{Major local (L2)}
      & secondary        & 12 & 0.36 \\
      & secondary\_link  & 12 & 0.27 \\
      & tertiary         & 12 & 0.30 \\
      & tertiary\_link   & 12 & 0.19 \\
    \midrule
    \multirow{3}{*}{Residential (L1)}
      & residential    & 4  & 0.10 \\
      & service        & 4  & 0.14 \\
      & unclassified   & 4  & 0.16 \\
    \midrule
    \multirow{8}{*}{Auxiliary}
      & pedestrian     & 4  & 0.34 \\
      & footway        & 4  & 0.14 \\
      & cycleway       & 4  & 0.055 \\
      & path           & 4  & 0.053 \\
      & steps          & 4  & 0.16 \\
      & track          & 4  & 0.014 \\
      & construction   & 2  & 0.029 \\
      & proposed       & 2  & 0.026 \\
    \bottomrule
  \end{tabular}
\end{table}

\subsubsection*{Training-time data augmentation.}

To compensate for the limited number of distinct training maps ($874$), each training image is augmented by the eight elements of the dihedral group of the square---identity, three $90^\circ$ rotations, horizontal and vertical flips, and the two diagonal reflections---yielding an eightfold increase in effective training pairs. Critically, every spatial token in the corresponding trajectory label is transformed by the matching symmetry operation, so that image and trajectory remain consistent across augmentations.

\subsection{Training Data Preparation}
\label{sec:experiments:training}

For each of the $874$ training meshes, we draw $100$ trajectories uniformly at random without replacement, producing $87{,}400$ distinct sampled training trajectories before augmentation. These trajectories serve as the ``captions'' of the corresponding map image. A batch element is one (map image, trajectory) pair; trajectories from the same mesh are distinct examples that share the conditioning image.

Equalizing the per-mesh sample count is essential: meshes differ by more than two orders of magnitude in their raw trajectory counts, so unbalanced pooling would let high-density meshes dominate the loss and encourage memorization of a few heavily represented (map, trajectory) associations. Capping at a common value makes the training signal a uniform average over the $874$ training meshes.

Each sampled trajectory is converted to a token sequence by applying the tokenization of Section~\ref{sec:method:token} pointwise to its $31.25\,\mathrm{m}$ cells, followed by the period token. Sequences are truncated to a fixed maximum length and zero-padded on the right for batching; padded positions contribute no gradient to the loss of Equation~\ref{eq:nll}. The loss is computed over all non-padding trajectory tokens, including the period token; prefixes used for dynamic continuation are supplied only at inference and excluded from the compared continuation segment.

Image augmentation under the dihedral group $D_4$ (Section~\ref{sec:experiments:maps}) is paired with trajectory-level augmentation: for every symmetry operation $\sigma$, the accompanying ground-truth trajectory is transformed by the same $\sigma$---its spatial-token sequence is relabeled by the induced action of $\sigma$ on the $32 \times 32$ cell grid, and its $(\varphi, \lambda)$ representation is rotated or reflected in sympathy---preserving geometric consistency within every augmented pair. The per-mesh cap of $100$, the eightfold augmentation, and the $874$-mesh training split yield $100 \times 8 \times 874 = 699{,}200$ effective image--trajectory pairs. The validation ($109$ meshes) and test ($110$ meshes) splits are prepared identically but without augmentation.

\paragraph{Data and implementation availability.}
OpenStreetMap road geometry and the Japanese standard regional-mesh definition are public. The Agoop mobility records used in this study cannot be redistributed by the authors under the data-use agreement. Agoop's Dynamic Population Data service is commercially available to prospective users upon request, subject to the provider's screening, licensing terms, and availability for the requested period and data fields; service details are provided in Ref.~\cite{Agoop2026}. The tokenization, OSM-rasterization, model-training, and evaluation code, together with the mesh split and bridge-edit specification, will be released upon publication.

\subsection{Evaluation Protocol}
\label{sec:experiments:eval}

Our evaluation addresses \emph{map sensitivity}, a necessary but not sufficient condition for editable-map simulation. For each mesh, generated trajectories are compared with held-out real trajectories under either the correct road raster or a shuffled raster from another mesh. The trajectory-level axis uses the same first two observed positions for both conditions; the density-level axis is map-only. The protocol therefore tests whether the specific map improves generation relative to an incorrect map. It does not measure deployment-grade absolute realism or establish superiority to external generators. Both axes are aggregated across meshes by a one-sided sign test.

\subsubsection*{Positive and negative generation sets.}

For every evaluation mesh $m$, we assemble three sets of trajectories of identical cardinality $n = 100$. The \emph{reference set} $R_m$ consists of $100$ real trajectories drawn from the mesh's pool of dynamic trajectories, disjoint from any trajectories used as training captions. The \emph{positive set} $G^+_m$ is generated under the correct OSM-derived map image $I_m$. To construct the negative set $G^-_m$, we first draw a random fixed-point-free permutation $\pi$ of the meshes within the same split, so that $\pi(m) \ne m$ for every evaluation mesh $m$, and then generate $100$ trajectories under the assigned OSM-derived map image $I_{\pi(m)}$. This shuffle-based negative control preserves the same cardinality as $G^+_m$ while testing whether the correct raster improves generation relative to a randomly assigned split-mate raster in the same mesh-local token frame. In the dynamic evaluation, positive and negative continuations share the same two target-mesh prefix tokens; only the raster is swapped.

\subsubsection*{Dynamic evaluation: trajectory-level energy distance.}

To quantify trajectory-level fidelity, we compare the distributions $R_m$ and $G_m \in \{G^+_m, G^-_m\}$ under a ground distance $d(\cdot, \cdot)$. We report two standard choices: the length-normalized Dynamic Time Warping (DTW) distance, which tolerates local temporal deformations, and the (bidirectional) Hausdorff distance, which is dominated by the worst-aligned point and is thus sensitive to isolated large geometric deviations. Both are computed in meters on the reconstructed coordinates from Equation~\ref{eq:decode}. The distribution-level discrepancy is the energy distance~\cite{Szekely2013},
\begin{equation}
\begin{split}
  D_E(R_m, G_m) = & \frac{2}{|R_m|\,|G_m|} \sum_{r \in R_m, g \in G_m} d(r, g) \\
  & - \frac{1}{|R_m|^2} \sum_{r, r' \in R_m} d(r, r') \\
  & - \frac{1}{|G_m|^2} \sum_{g, g' \in G_m} d(g, g'),
\end{split}
\label{eq:energy}
\end{equation}
which we use as a distributional discrepancy induced by the chosen trajectory distance. Because DTW and Hausdorff are used as trajectory distances rather than Euclidean vector distances, our inference relies on within-mesh positive--negative differences, not on the strict identification property of classical energy distance. The per-mesh improvement of positive over negative is
\begin{equation}
  \Delta D_{E,m}
  \;=\;
  D_E(R_m,\, G^-_m) \;-\; D_E(R_m,\, G^+_m),
  \label{eq:deltaDE}
\end{equation}
so that $\Delta D_{E,m} > 0$ means the correct map brings generated trajectories closer to reality in distribution than the mismatched map does.

\subsubsection*{Static evaluation: density-level heatmap correlation.}

Trajectory-level fidelity does not capture where \emph{populations} of trajectories concentrate. For each $X \in \{R_m, G^+_m, G^-_m\}$, we rasterize its $100$ trajectories onto the $224 \times 224$ map grid to obtain a heatmap $H_X \in \mathbb{R}^{224 \times 224}$ whose $(u, v)$ entry records the number of trajectories passing through pixel $(u, v)$, and compute the pixelwise Pearson correlation $\mathrm{corr}(H_{R_m}, H_{G_m})$. The per-mesh improvement is
\begin{equation}
  \Delta \mathrm{corr}_m
  \;=\;
  \mathrm{corr}(H_{R_m},\, H_{G^+_m})
  \;-\;
  \mathrm{corr}(H_{R_m},\, H_{G^-_m}),
  \label{eq:deltacorr}
\end{equation}
with $\Delta \mathrm{corr}_m > 0$ indicating that the correct map yields a density closer to the observed one than the mismatched map does.

\subsubsection*{Mesh-level aggregation by sign test.}

Two properties of the evaluation make a naive mean across meshes unreliable: the scale of $\Delta D_{E,m}$ varies by orders of magnitude due to heterogeneity in trajectory length and road density, so a few outliers can dominate any average; and the quantity of primary interest is \emph{consistency}, not magnitude---whether the correct map outperforms the mismatched one on most meshes. We therefore aggregate by a one-sided sign test: for each configuration, ground distance, and split, we report the mesh-level positive ratio $\hat p = \#\{m : \Delta_m > 0\} / M$ (where $\Delta_m$ is the relevant per-mesh statistic and $M$ is the number of meshes in the split) together with the exact one-sided binomial $p$-value against $H_0 : \Pr(\Delta_m > 0) \le \tfrac{1}{2}$.

\section{Results}
\label{sec:results}

\subsection{Training Convergence}
\label{sec:results:training}

\begin{figure}[t]
  \centering
  \begin{tabular}{@{}c@{}}
    \includegraphics[width=0.75\linewidth]{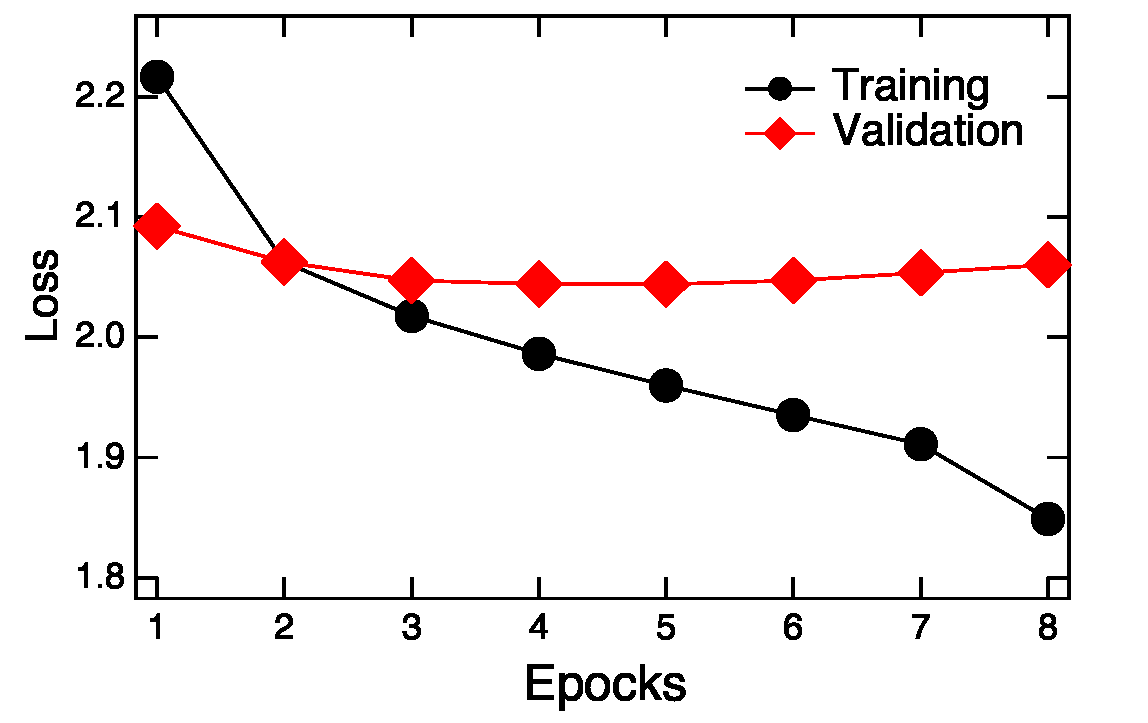} \\
    \small (a) CNN-based configuration \\[4pt]
    \includegraphics[width=0.75\linewidth]{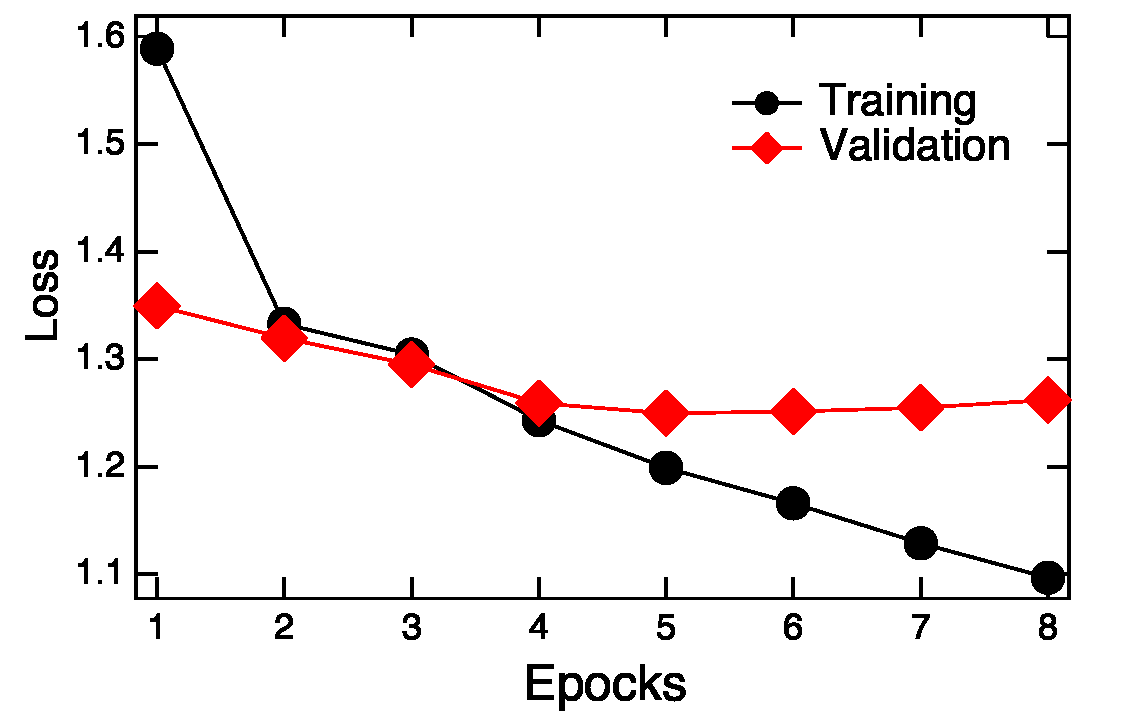} \\
    \small (b) ViT-based configuration
  \end{tabular}
  \caption{Training and validation loss. The selected checkpoints are epoch~$4$ for the CNN configuration and epoch~$5$ for the ViT configuration.}
  \Description{Two plots show training and validation loss over eight epochs for CNN- and ViT-conditioned trajectory generators.}
  \label{fig:training}
\end{figure}

Both variants were trained end-to-end from random initialization for eight epochs using the token-level negative log-likelihood of Equation~\ref{eq:nll}. Each epoch traverses the $699{,}200$ geometry-consistent augmented image--trajectory pairs derived from $87{,}400$ sampled trajectories and $874$ distinct maps; the training set is therefore not limited to $100$ trajectories in total. Figure~\ref{fig:training} shows the learning curves. The CNN validation-loss minimum occurs at epoch~$4$ and the ViT's at epoch~$5$; these checkpoints are adopted for all subsequent evaluations (Sections~\ref{sec:results:qualitative}--\ref{sec:results:static}). Notably, the ViT attains a substantially lower validation loss than the CNN (approximately $1.2$ versus $2.0$, corresponding to per-token perplexities of $3.3$ and $7.4$): on the proxy metric that the model is directly optimized for---the probability of the next spatial token---the Vision Transformer fits the training distribution more tightly. As will become clear in Sections~\ref{sec:results:dynamic}--\ref{sec:results:static}, this training-set advantage does not transfer to geometric trajectory quality, and the dissociation is itself informative.

\subsection{Generation Settings and Qualitative Analysis}
\label{sec:results:qualitative}

\begin{figure}[t]
  \centering
  \begin{tabular}{@{}c@{}}
    \includegraphics[width=\linewidth]{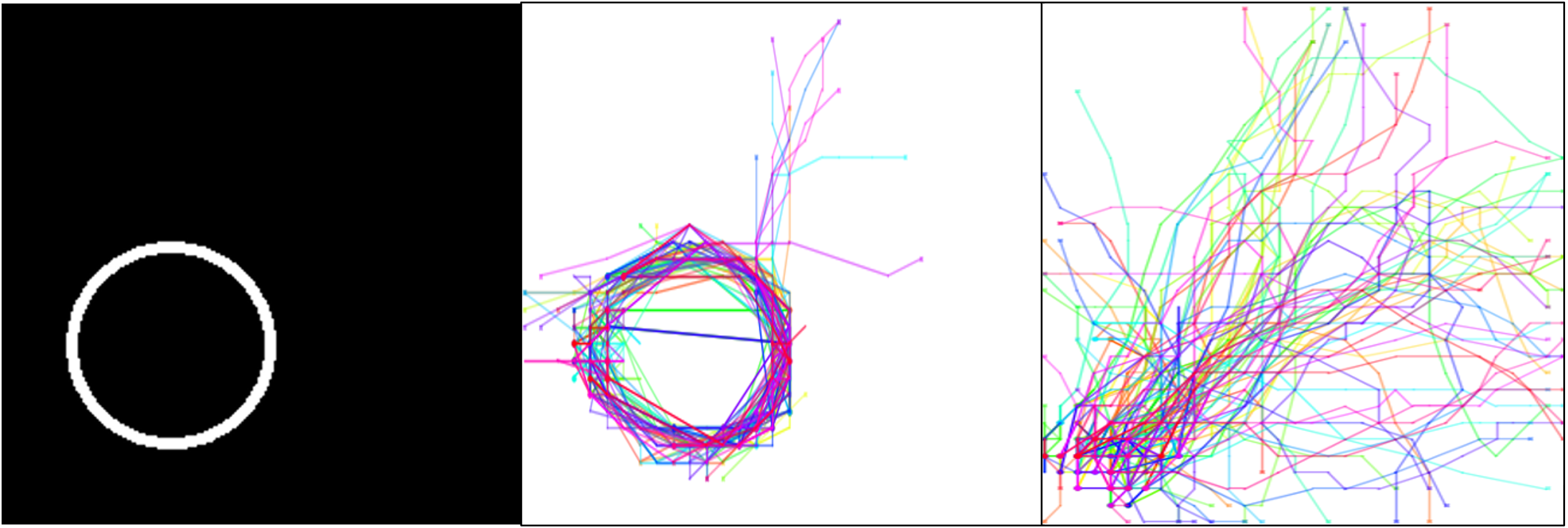} \\
    \small (a) Synthetic circular road \\[4pt]
    \includegraphics[width=\linewidth]{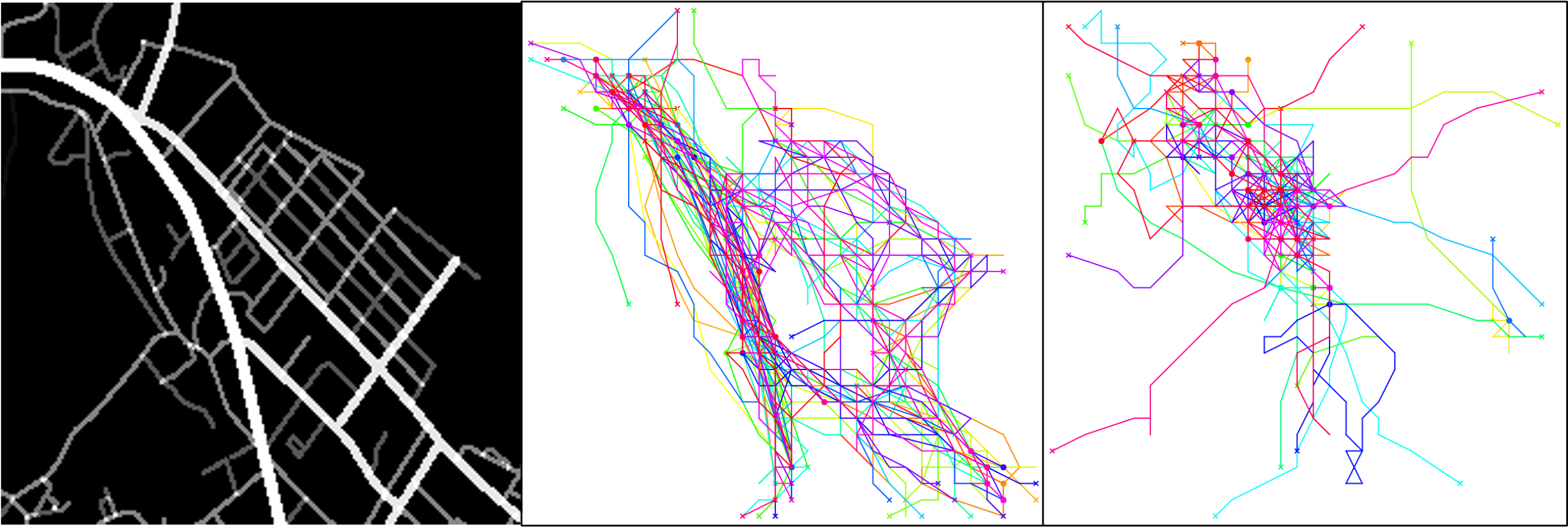} \\
    \small (b) OSM-derived urban road image
  \end{tabular}
  \caption{Map-only comparison of $100$ trajectories per configuration. From left: conditioning map, CNN generations, and ViT generations. (a) Synthetic circular road. (b) OSM-derived urban road image.}
\Description{Two rows compare a conditioning map with one hundred CNN- and ViT-generated trajectories on a circular road and an urban road raster.}
\label{fig:qualitative}
\end{figure}

Generation proceeds autoregressively from the checkpoints of Section~\ref{sec:results:training}. We use two sampling modes. Figure~\ref{fig:qualitative} and the heatmap evaluation are map-only: no position token is supplied, and the first spatial token is sampled from the map condition. Dynamic trajectory evaluation and the bridge intervention condition only on the first two observed positions, so continuations share the same initial movement. In both modes, subsequent tokens are drawn with $\mathrm{top\text{-}}k = 50$, $\mathrm{top\text{-}}p = 0.9$, and temperature $1.0$ until the period token or maximum length is reached.

Figure~\ref{fig:qualitative} shows $100$ map-only trajectories for each configuration on (a) a synthetic circular road and (b) an OSM-derived urban mesh. The examples are visual sanity checks, not quantitative evidence. Because the training raster is built from human-mobility samples rather than sidewalk polygons, observed positions may occur beside rendered OSM centerlines at sidewalks, entrances, or adjacent lots. We therefore do not treat strict road-centerline adherence as a correctness criterion.

The CNN traces the circular network in both directions and concentrates generations near visible corridors in the OSM example, with occasional larger deviations. The ViT localizes activity near the dominant corridor but preserves local road geometry less consistently, including paths through the interior of the synthetic ring. This contrast motivates the trajectory- and density-level map-sensitivity tests below.

\subsection{Trajectory-Level Map-Sensitivity Evaluation}
\label{sec:results:dynamic}

\begin{figure*}[t]
  \centering
  \begin{tabular}{@{}cc@{}}
    \includegraphics[width=0.35\linewidth]{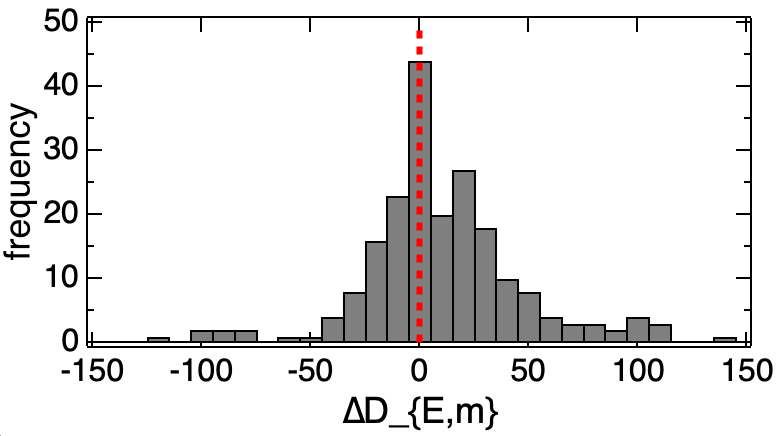} &
    \includegraphics[width=0.35\linewidth]{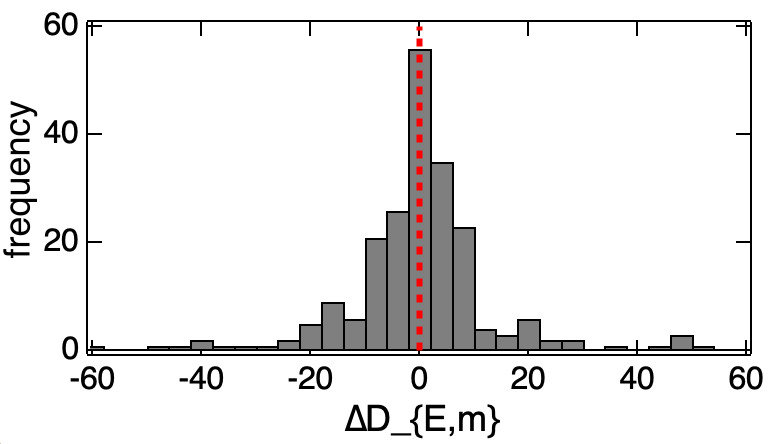} \\
    \small (a) CNN, Hausdorff &
    \small (b) ViT, Hausdorff \\[6pt]
    \includegraphics[width=0.35\linewidth]{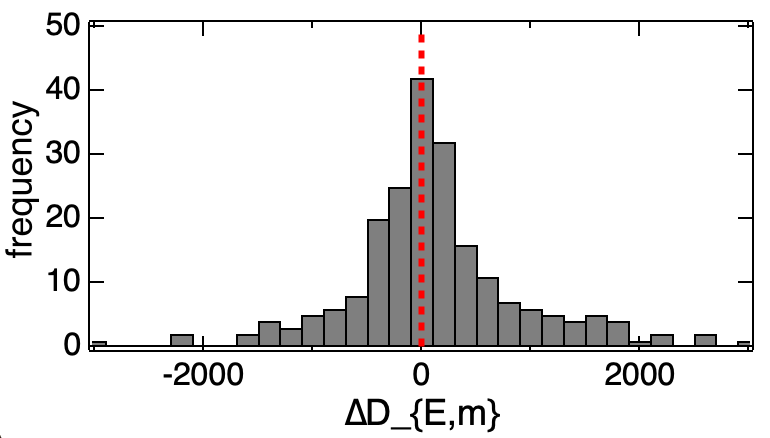} &
    \includegraphics[width=0.35\linewidth]{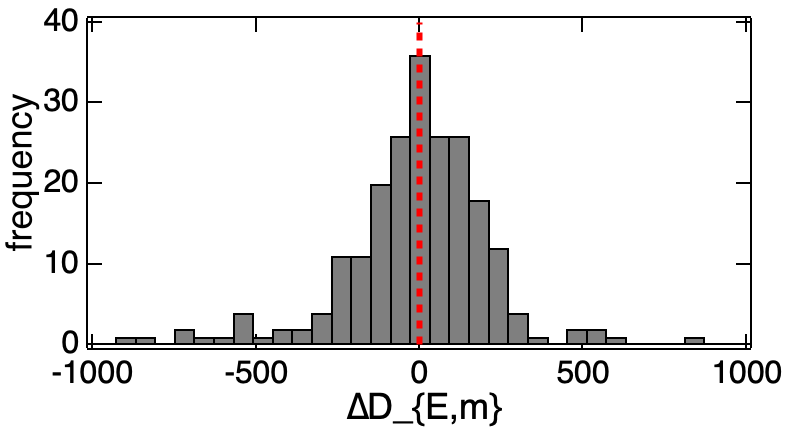} \\
    \small (c) CNN, DTW &
    \small (d) ViT, DTW \\
  \end{tabular}
  \caption{Per-mesh improvement $\Delta D_{E,m}$ on $219$ held-out meshes. Rows: Hausdorff and DTW; columns: CNN and ViT. Positive values favor the correct map over the shuffled map.}
\Description{Four histograms show correct-map improvement for CNN and ViT models under Hausdorff and DTW trajectory distances.}
\label{fig:dynamic}
\end{figure*}

Following Section~\ref{sec:experiments:eval}, we evaluate $\Delta D_{E,m}$ with two trajectory distances: length-normalized DTW and Hausdorff. Table~\ref{tab:dynamic} reports the positive ratio $\hat p$ and one-sided binomial $p$-value across splits. Figure~\ref{fig:dynamic} shows the full $\Delta D_{E,m}$ distributions for the $219$ validation and test meshes, none of whose maps appeared in training. The CNN distributions are shifted to the right, especially under Hausdorff, indicating that the correct map helps on a majority of held-out meshes. The ViT distributions remain closer to zero, indicating weaker trajectory-level map sensitivity.

\begin{table}[t]
  \centering
  \small
  \caption{Trajectory-level map sensitivity. Positive ratio $\hat p$ and one-sided binomial $p$-value for $\Pr(\Delta D_{E,m}>0)>1/2$. Values above $0.5$ favor the correct map on a majority of meshes. Because the $p$-value is a deterministic function of the number of positive meshes, the identical DTW test entries for CNN and ViT ($\hat p = 0.57$, $p = 0.074$) reflect the same count of positive meshes and are not a transcription error.}
  \label{tab:dynamic}
  \begin{tabular}{llcccc}
    \toprule
    & & \multicolumn{2}{c}{CNN} & \multicolumn{2}{c}{ViT} \\
    \cmidrule(lr){3-4}\cmidrule(lr){5-6}
    Distance & Split & $\hat p$ & $p$-value & $\hat p$ & $p$-value \\
    \midrule
    \multirow{3}{*}{Hausdorff}
      & Train      & $0.71$ & $10^{-38}$ & $0.47$ & $0.944$ \\
      & Validation & $0.62$ & $0.0079$   & $0.44$ & $0.895$ \\
      & Test       & $0.60$ & $0.021$   & $0.50$ & $0.538$ \\
    \midrule
    \multirow{3}{*}{DTW}
      & Train      & $0.62$ & $10^{-12}$ & $0.49$ & $0.740$ \\
      & Validation & $0.58$ & $0.051$    & $0.47$ & $0.750$ \\
      & Test       & $0.57$ & $0.074$    & $0.57$ & $0.074$ \\
    \bottomrule
  \end{tabular}
\end{table}

The two configurations differ sharply, especially under the Hausdorff distance. For the CNN variant, Hausdorff-based positive ratios lie between $0.60$ and $0.71$ across all three splits, with the one-sided sign test rejecting $H_0$ at $p < 10^{-2}$ on the training and validation splits; on the held-out test split of $110$ meshes whose map images were not seen during training, the CNN retains $\hat p = 0.60$ at $p = 0.021$. This supports the claim that the correct road raster improves trajectory-level fidelity on unseen road layouts relative to the shuffled-map control. Under DTW, the CNN variant also retains a positive-ratio majority across splits ($0.57$ to $0.62$), with the training split clearly significant ($p = 10^{-12}$) and the validation and test splits borderline ($p = 0.051$ and $0.074$, respectively). We therefore interpret the DTW result as directional but not conclusive on the held-out splits. The ViT variant, in contrast, produces positive ratios close to $\tfrac{1}{2}$ on most split--distance combinations and does not attain conventional significance. Taken together with the qualitative examples of Section~\ref{sec:results:qualitative}, these results suggest that the ViT-based configuration transmits enough global information to localize activity but provides less reliable local structure for stepwise trajectory geometry. Because the shuffled negative rasters are themselves valid OSM-derived road networks with the same class of geometric and statistical features, the CNN configuration's correct-map gain is not explained by generic road-image priors alone; it depends on the specific road geometry of the evaluation mesh. The contrast with the training-loss ordering of Section~\ref{sec:results:training} is striking: the ViT wins on the metric directly optimized during training, yet the CNN is stronger on geometric trajectory metrics, an inversion we return to in Section~\ref{sec:results:discussion}.

\subsection{Density-Level Map-Sensitivity Evaluation}
\label{sec:results:static}

Following the protocol of Section~\ref{sec:experiments:eval}, we aggregate the per-mesh improvement $\Delta \mathrm{corr}_m$ by the one-sided sign test. Table~\ref{tab:static} reports the mean reference--generated correlations $\overline{\mathrm{corr}}^{+}$ and $\overline{\mathrm{corr}}^{-}$ alongside $\hat p$ and the one-sided binomial $p$-value across splits.

Both configurations reject $H_0$ on every split, but at quantitatively different levels: the CNN variant exceeds $\hat p = 0.94$ throughout, while the ViT variant remains in the $0.75$--$0.79$ range. On the test split, the CNN reaches $\overline{\mathrm{corr}}^{+} = 0.38$ against $\overline{\mathrm{corr}}^{-} = 0.01$, showing that the correct road raster recovers the observed spatial concentration of movement on held-out meshes. The ViT achieves a smaller but nontrivial $\overline{\mathrm{corr}}^{+} = 0.14$ with $\overline{\mathrm{corr}}^{-} = 0.00$. This static result is important but should be interpreted separately from dynamic trajectory fidelity: matching where people tend to concentrate is easier than reproducing the geometry of individual stepwise paths. The static--dynamic gap is precisely why we report both axes.

\begin{table}[t]
  \centering
  \footnotesize
  \setlength{\tabcolsep}{4pt}
  \caption{Density-level map sensitivity. Mean heatmap correlations for correct ($\overline{\mathrm{corr}}^{+}$) and shuffled ($\overline{\mathrm{corr}}^{-}$) maps, positive ratio $\hat p$, and one-sided binomial $p$-value.}
  \label{tab:static}
  \begin{tabular}{lcccc|cccc}
    \toprule
    & \multicolumn{4}{c|}{CNN} & \multicolumn{4}{c}{ViT} \\
    \cmidrule(lr){2-5}\cmidrule(lr){6-9}
    Split & $\overline{\mathrm{corr}}^{+}$ & $\overline{\mathrm{corr}}^{-}$ & $\hat p$ & $p$-value & $\overline{\mathrm{corr}}^{+}$ & $\overline{\mathrm{corr}}^{-}$ & $\hat p$ & $p$-value \\
    \midrule
    Train      & 0.41 & 0.00 & $0.97$ & $10^{-214}$ & 0.15 & 0.00 & $0.79$ & $10^{-69}$ \\
    Validation & 0.41 & 0.03 & $0.96$ & $10^{-26}$  & 0.13 & 0.01 & $0.77$ & $10^{-9}$  \\
    Test       & 0.38 & 0.01 & $0.94$ & $10^{-13}$  & 0.14 & 0.00 & $0.75$ & $10^{-8}$ \\
    \bottomrule
  \end{tabular}
\end{table}

\subsection{Editable-Map Intervention Example}
\label{sec:results:counterfactual}

The preceding evaluation tested map sensitivity on held-out but unedited meshes. We now show an illustrative intervention in which only a localized part of the input raster is changed. This is an interface demonstration, not a quantitative test of causal or planning validity.

\begin{figure}[t]
  \centering
  \includegraphics[width=0.9\linewidth]{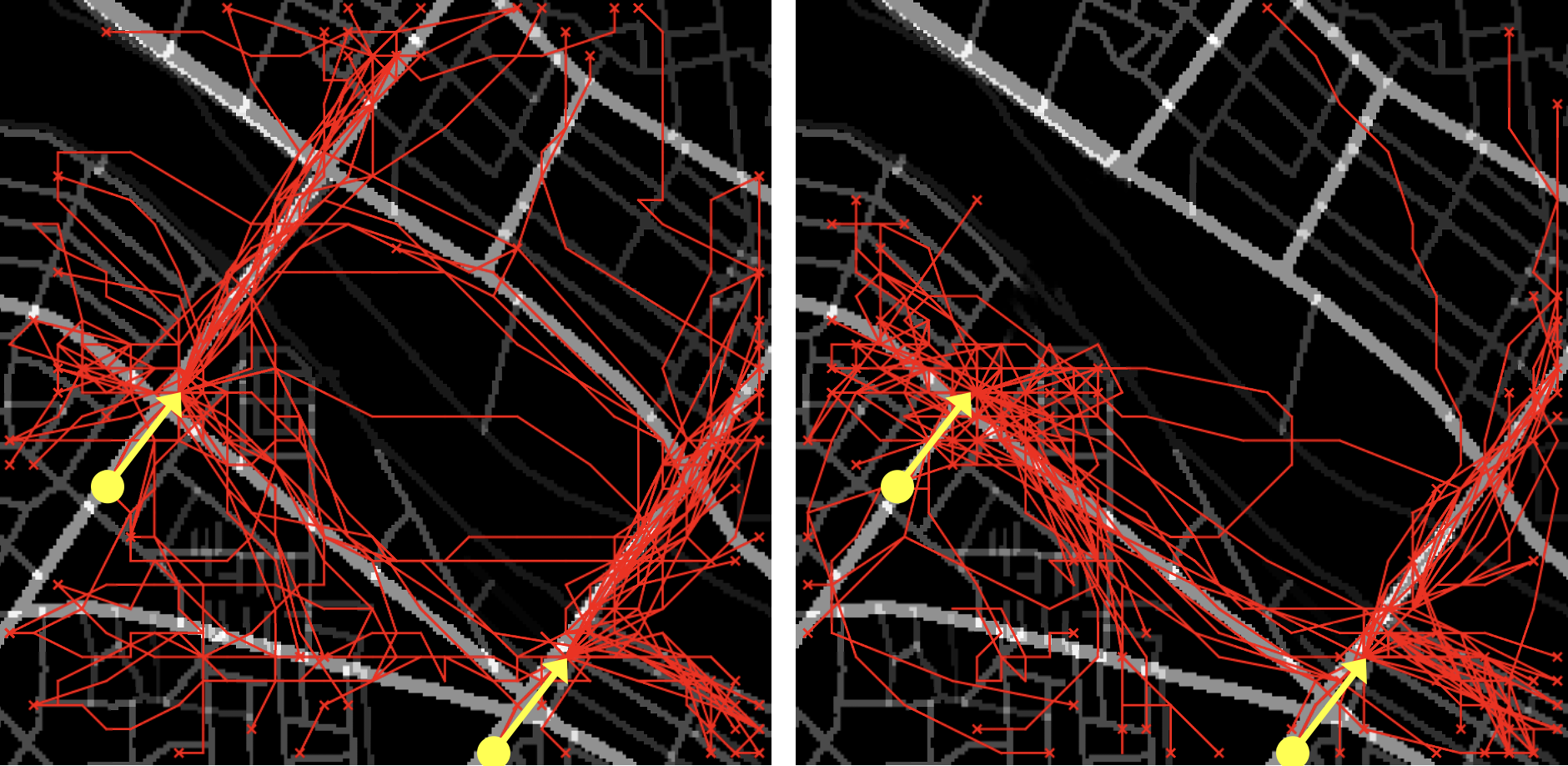}
  \caption{Illustrative edited-map intervention. CNN-generated continuations (red; $50$ from each of two prefixes) under the original map (left) and a map with the northwestern bridge removed (right). Only the input raster changes. Both panels show a single $1\,\mathrm{km}$ mesh; generated positions are the centroids of its $32\times32$ mesh-local cells (nominal $31.25\,\mathrm{m}$), which is why the paths appear much finer than the mesh itself.}
\Description{Side-by-side maps show generated trajectories before and after one bridge is removed from the road raster.}
\label{fig:counterfactual}
\end{figure}

Figure~\ref{fig:counterfactual} shows $100$ CNN-generated continuations ($50$ from each of two two-position prefixes) on a test mesh with two arterial bridges across a river. The left panel uses the original OSM-derived raster. In the right panel, the northwestern bridge is removed while the model, sampling settings, and prefixes remain unchanged. Visual inspection shows less generated flow across the removed connection, accumulation near its approach, and some redistribution toward the remaining bridge. These observations are qualitative; no bridge-crossing or detour-rate statistic is claimed.

The example demonstrates the operational mechanism relevant to geospatial simulation: edit the map, hold the model and initial state fixed, and regenerate the continuation. A broader perturbation set and quantitative measures of crossing, rerouting, and displacement are required before the outputs can be used for planning decisions.

\subsection{Discussion}
\label{sec:results:discussion}

The encoder comparison is a secondary empirical finding within the editable-map interface. Under the present data scale and from-scratch training regime, the CNN-based configuration is more sensitive to the specific local road raster than the ViT-based configuration. One possible explanation is that local receptive fields bias the CNN toward nearby segments and junctions, whereas a ViT must learn such locality from the data~\cite{Raglu2021}. However, encoder family is confounded with conditioning pathway: the CNN uses a visual prefix, whereas the ViT uses cross-attention. The present experiment therefore cannot isolate whether the observed gap arises from encoder locality, the conditioning mechanism, or their interaction. Separating these effects requires a crossed ablation that holds the conditioning pathway fixed across encoder families.

The loss--geometry mismatch is also informative. The ViT attains lower token-level negative log-likelihood, yet its generated trajectories are weaker under map-sensitive geometric metrics. Cross-entropy treats every incorrect next token categorically, whereas trajectory distances distinguish small local errors from isolated large geometric deviations. Training a spatial sequence model is therefore not sufficient by itself; its outputs must also be evaluated in geographic space.

The scale should be stated precisely. Training uses $874$ distinct maps and $87{,}400$ sampled real trajectories. Geometry-preserving augmentation produces $699{,}200$ image--trajectory pairs. The number of maps remains modest for a from-scratch visual encoder and may particularly disadvantage the ViT, but the experiment is not based on only $100$ trajectories in total.

The shuffled-map protocol answers one specific question: does the correct map matter relative to an incorrect map when both outputs are compared with held-out real trajectories? It does not establish absolute realism, utility for planning, or superiority to coordinate-, diffusion-, imitation-learning, or rule-based simulators. Because the data represent human movement across modes and include locations beside road centerlines, future realism tests should compare generated and real road-proximity distributions rather than maximize a vehicle-style road-adherence rate. They should also include trip-length, displacement, duration, and kinematic statistics such as speed and acceleration, together with quantitative intervention metrics, external baselines, and additional regions.

\section{Conclusion and Future Work}
\label{sec:conclusion}

We presented an editable map-conditioned interface for human-mobility simulation. A road raster conditions an autoregressive decoder that emits spatial-token trajectories at one-minute resolution on a mesh-local grid, allowing the same model and vocabulary to be applied to held-out or locally edited maps. The main contribution is this geospatial simulation interface, not a new CNN, ViT, or tokenization primitive.

On $1{,}093$ meshes in Ishikawa Prefecture, correct-map generations were compared with shuffled-map generations against held-out real trajectories. The CNN-conditioned model showed the clearest map sensitivity under Hausdorff-based energy distance and density-level heatmap correlation; DTW evidence was weaker, and the ViT was less reliable at trajectory-level grounding. A bridge-removal example further illustrated how a local raster edit can alter generated continuations without retraining.

These results establish feasibility rather than deployment-grade realism. Future work will compare external trajectory generators, evaluate absolute movement statistics and road-proximity distributions, quantify responses over a benchmark of map perturbations, and test additional regions and pretrained visual encoders.

\begin{acks}
This work was supported by JSPS KAKENHI JP25K01453 (T.M.), JP25K01458 (T.M.), JP25K08191 (A.I.), and JP26K01163 (S.F., A.I., T.M.), and by MEXT Supporting Pioneering Research through AI for 1,000 Discovery Challenges Program (SPReAD), Japan, Grant Number JPMXP1726298309 (S.F.). We thank Agoop Corporation for the Dynamic Population Data. ChatGPT was used only for English polishing; the authors reviewed and take responsibility for all scientific content, analyses, citations, and interpretations.
\end{acks}

\end{document}